\documentclass[letterpaper, 10 pt, conference]{ieeeconf}  

\IEEEoverridecommandlockouts                              

\usepackage{url}
\usepackage{hyperref}
\hypersetup{
    colorlinks=true,       
    linkcolor=magenta,     
    urlcolor=magenta       
}

\usepackage[tracking=true]{microtype}
\usepackage{amsmath} 
\usepackage{amssymb}  
\usepackage{graphicx}
\usepackage{epsfig} 
\usepackage{mathptmx} 
\usepackage{times} 
\usepackage{amsmath} 
\usepackage{amssymb}  
\usepackage{xspace}

\usepackage{listings}
\usepackage{float}
\usepackage{afterpage}
\usepackage{url}
\usepackage[dvipsnames]{xcolor}
\usepackage[font={footnotesize}]{caption}
\usepackage{booktabs} 
\usepackage{tikz}
\usetikzlibrary{positioning,shapes.geometric,arrows,fit,shapes.symbols,svg.path,tikzmark,calc}
\usepackage{textcomp}
\usepackage{listings}
\usepackage{subcaption}
\pgfdeclarelayer{background}
\pgfdeclarelayer{foreground}
\pgfsetlayers{background,main,foreground}
\usepackage{siunitx}
\usepackage{adjustbox}
\usepackage{array}
\usepackage{multirow}
\usepackage{hyperref}
\usepackage{siunitx}

\let\labelindent\relax 

\usepackage[inline]{enumitem} 

\usepackage[backend=biber,
            url=false,
            isbn=false,
            doi=false,
            backref=false,
            style=ieee,
            natbib=true,
            mincitenames=1,
            maxcitenames=1,
            citestyle=numeric-verb,
            sorting=none,
            block=none]{biblatex}
\renewcommand{\bibfont}{\small}
\title{\LARGE \bf
Language-Embedded Gaussian Splats (LEGS): \\ Incrementally Building Room-Scale Representations with a Mobile Robot
}

\author{Justin Yu*$^{1}$, Kush Hari*$^{1}$, Kishore Srinivas*$^{1}$, 
Karim El-Refai$^{1}$, Adam Rashid$^{1}$, Chung Min Kim$^{1}$, Justin Kerr$^{1}$, \\Richard Cheng$^{2}$, Muhammad Zubair Irshad$^{2}$, Ashwin Balakrishna$^{2}$, Thomas Kollar$^{2}$, Ken Goldberg$^{1}$ 
\thanks{$^{*}$ Equal contribution}%
\vspace{0.4cm} \\
\url{berkeleyautomation.github.io/LEGS}
\vspace{-0.2cm}
\thanks{$^{1}$The AUTOLab at UC Berkeley (automation.berkeley.edu).}
\thanks{$^{2}$Toyota Research Institute, Los Altos, CA.}
}

\begin{document}

\maketitle

\begin{abstract}
    Building semantic 3D maps is valuable for searching for objects of interest in offices, warehouses, stores, and homes. We present a mapping system that incrementally builds a Language-Embedded Gaussian Splat (LEGS): a detailed 3D scene representation that encodes both appearance and semantics in a unified representation.
LEGS is trained online as a robot traverses its environment to enable localization of open-vocabulary object queries.
We evaluate LEGS on 4 room-scale scenes where we query for objects in the scene to assess how LEGS can capture semantic meaning. We compare LEGS to LERF \cite{kerr2023lerf} and find that while both systems have comparable object query success rates, LEGS trains over 3.5x faster than LERF. 
Results suggest that a multi-camera setup and incremental bundle adjustment can boost visual reconstruction quality in constrained robot trajectories, and suggest LEGS can localize open-vocabulary and long-tail object queries with up to 66\% accuracy. 
\end{abstract}

\section{Introduction}
Consider open vocabulary robot requests such as \textit{``Where are gluten-free crackers?"} or \textit{``Get a stain remover spray"}, a robot must parse such queries, localize relevant objects, and navigate to them. A large body of recent work uses large vision-language models by distilling their outputs into 3D representations like point clouds or NeRFs \cite{mildenhall2021nerf}. These semantic representations have been applied to both manipulation~\cite{ze2023gnfactor,shen2023F3RM,lerftogo2023} and large-scale scene understanding~\cite{conceptfusion,shafiullah2023clipfields,huang23vlmaps}, showing promise of using large models zero-shot for open-vocabulary task specification.

One key challenge for scaling these methods to large environments is the underlying 3D representation, which should be flexible to a variety of scales, able to update with new observations, and fast. Although NeRFs are commonly used as the 3D representation for distilling 2D semantic features~\cite{kobayashi2022decomposing,tschernezki2022neural,kerr2023lerf}, scaling NeRFs to large scenes can be cumbersome because 
they typically rely on a fixed spatial resolution
~\cite{meuleman2023progressively,wang2023f2,tancik2022block}, are difficult to modify, and slower to render.
A popular alternative is pointclouds~\cite{peng2023openscene,shafiullah2023clipfields, huang23vlmaps, conceptfusion}, which work seamlessly with many SLAM algorithms. 
However, a given point is assigned a single color and semantic feature by fusing CLIP in the pointcloud with a contrastively supervised field, whereas a multi-scale model of the world can simultaneously reason about objects and their parts, similar to how LERF-TOGO~\cite{lerftogo2023} leverages multi-scale semantics in LERF~\cite{kerr2023lerf}.  

\begin{figure}
    \centering
    \includegraphics[width=\linewidth]
    {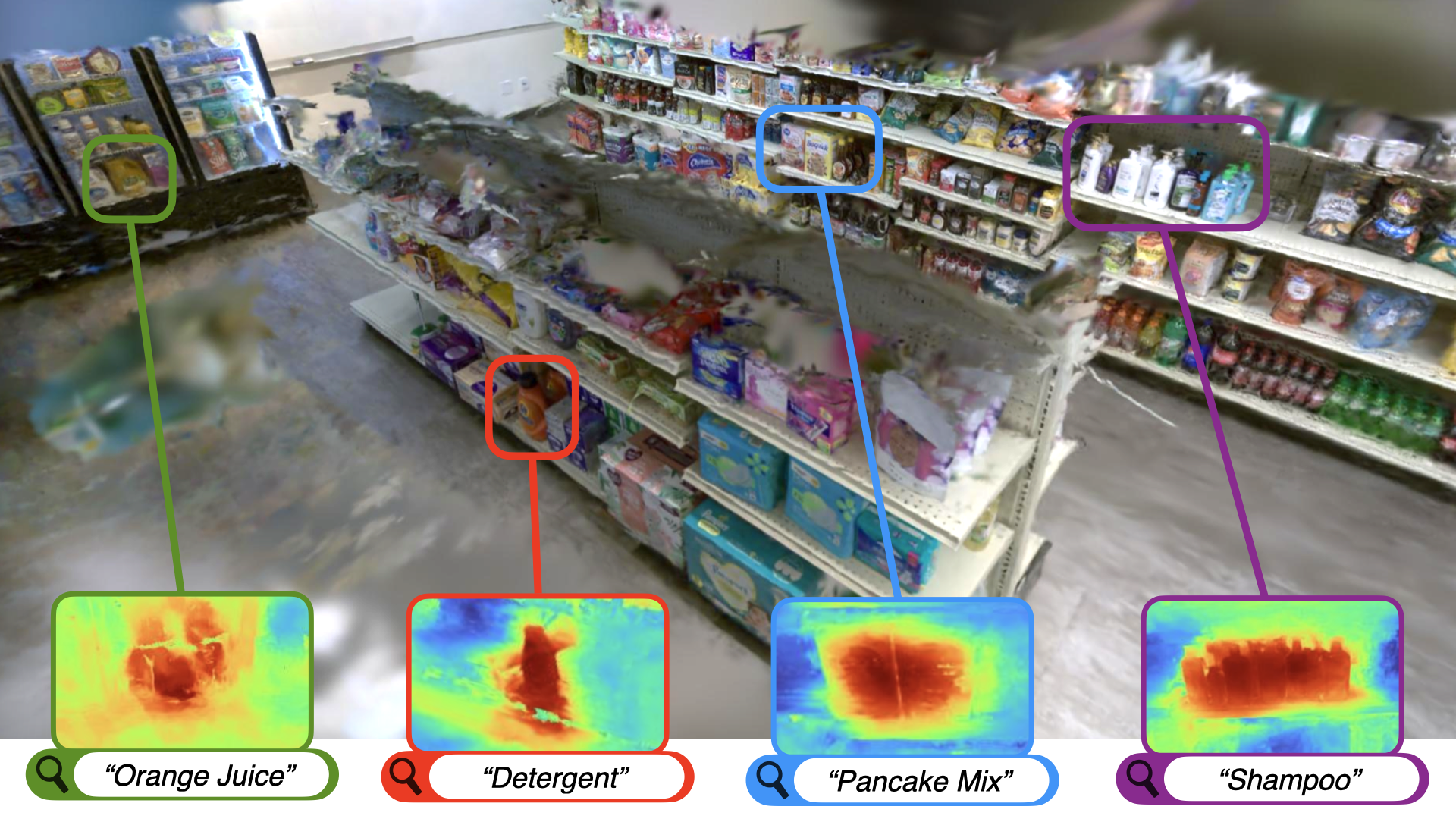}
    \caption{\textbf{Language-Embedded Gaussian Splat in TRI Grocery Store Testbed \cite{bajracharya2024demonstrating}.} LEGS relies entirely on pretrained VLMs and does not require any inventory data or finetuning.}
    \label{fig:splash}
    \centering
    \vspace*{-0.3in}
\end{figure}
3D Gaussian Splatting (3DGS)~\cite{kerbl20233d} models the 3D scene using a large set of 3D Gaussians. Recent works \cite{zuo2024fmgs, qin2023langsplat} successfully assign semantic features to every Gaussian in the scene. However, existing techniques combining semantic features and 3D Gaussian Splatting (3DGS) scene reconstruction require offline computation of keyframe transforms and 3D Gaussian initialization points. 


% ...

In this paper, we focus on linking language understanding to Gaussian Splats in large-scale scenes, while incrementally training on a stream of RGBD images of the scene from a mobile robot. This incremental training method offers substantial benefits, notably enabling the robot to autonomously determine its position within the environment and subsequently use the map data for enhanced operational efficiency.



LEGS combines geometry and appearance information from 3DGS with semantic knowledge from CLIP by grounding language embeddings into the 3DGS similar to the method described in \cite{zuo2024fmgs}. LEGS incrementally registers images and simultaneously optimizes both 3D Gaussians and dense language fields. This allows robots to build maps that contain rich representations of their surroundings that can be queried with natural language.

This paper makes 3 contributions:
\begin{itemize}
    
    \item An online multi-camera 3DGS reconstruction system for large-scale scenes. The system takes as input three video streams from a mobile robot, and incrementally builds the 3D scene.
    
    \item Language-Embedded Gaussian Splatting (LEGS), a hybrid 3D semantic representation that uses explicit 3D Gaussians for geometry and implicit scale-conditioned hashgrid~\cite{mueller2022instant} for the semantics. 
    
    \item Results from physical experiments suggesting LEGS can produce high quality Gaussian Splats in room-scale scenes with training time 3.5x faster than a LERF baseline~\cite{kerr2023lerf}. 
\end{itemize}


\section{Related work}
\begin{figure*}[!t]
\centering
\includegraphics[width=0.8\linewidth]{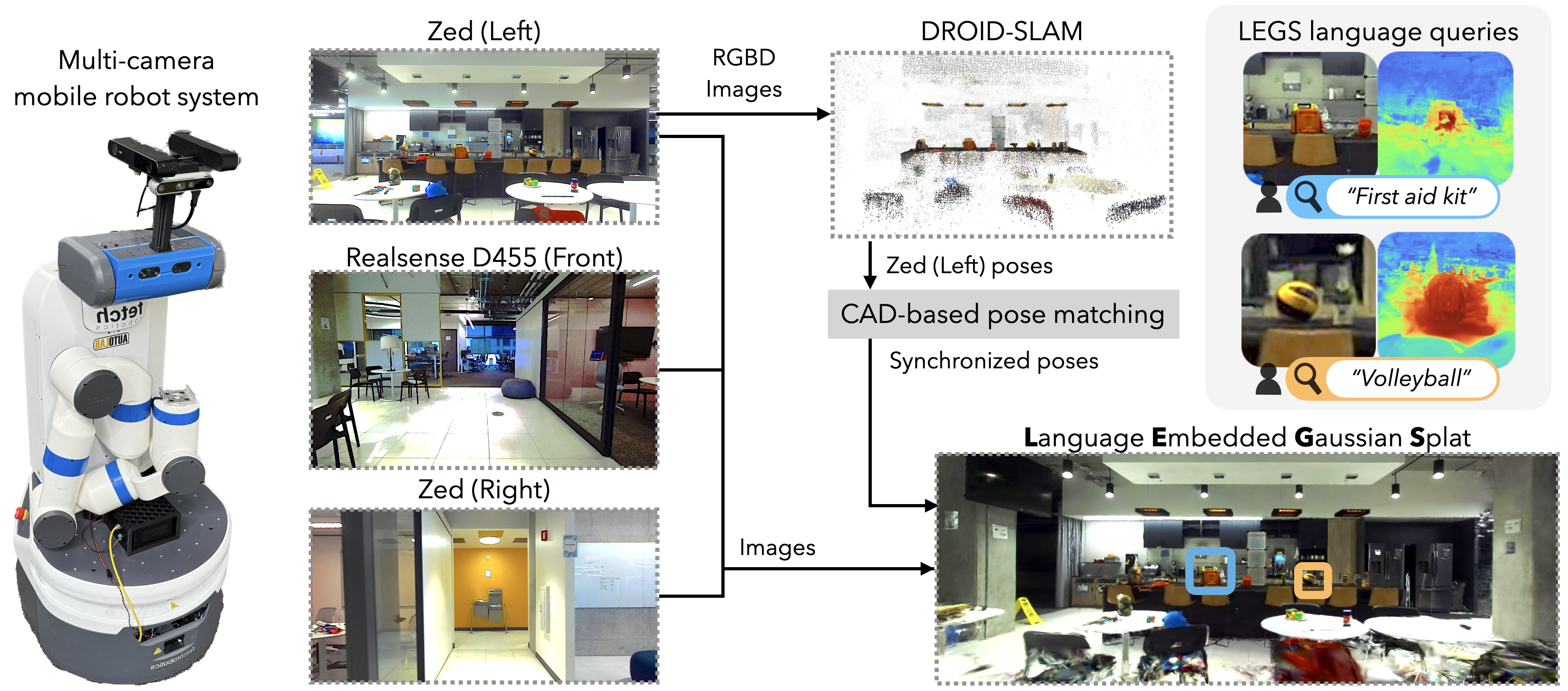}
\caption{\textbf{LEGS System Integration} For LEGS, we use a Fetch robot with a custom multicamera configuration where a Realsense D455 is facing forward while 2 Zed cameras face the left and right sides respectively. The left Zed image stream is inputted into DROID-SLAM to compute pose estimates for the left camera, and the corresponding extrinsics are used to compute the pose estimates for the other Zed camera and D455. These image-poses are then used for concurrent Gaussian splat and CLIP training online. From there, the Gaussian splat can be queried for an object (ex. ``First Aid Kit"), and the corresponding relevancy field will be computed to localize the desired object.}
\label{fig:system_integration}
\vspace*{-0.3in}
\end{figure*}




\subsection{Mobile Robot Mapping}
Early robotic scene mapping research focused on the development of the core competencies in the metric \cite{arras2003feature,chatila1985position,jiang2019simultaneous} and topological \cite{choset2001topological,tapus2005topological} knowledge spaces, extensively centered around the question of map and knowledge representation. For successful task execution, data-rich 3D scene representation and self-localization are critical, enabled by Simultaneous Localization and Mapping (SLAM) algorithms \cite{alsadik2021simultaneous,kohlbrecher2011flexible,hess2016real}. 3D spatial maps have traditionally been represented by voxel grids, points or surfels, and more recently, neural radiance fields \cite{huang2021review,lazaro2018efficient}. Each of these approaches come with their own limitations. The accuracy and expressiveness of occupancy and voxel grids are resolution-bounded due to quantization. Points and surfels are discontinuous when rendered, making it challenging to supervise features in a continuous manner. Recent SLAM methods adopting a neural radiance field representation such as NICE-SLAM \cite{zhu2022nice} and NeRF-SLAM \cite{rosinol2023nerf} are constrained by their implicit representation making it difficult to update geometry over time.

\subsection{Semantic Scene Mapping for Robotics}

Semantic grounding, particularly in 3D representations, is a longstanding problem \cite{roldao20223d} to integrate semantic knowledge of objects and the surrounding environment into a mapped scene. The first definition of semantic mapping for robotics is provided by Nüchter, \textit{et al.} as a spatial map, 2D or 3D, augmented by information about entities, i.e., objects, functionalities, or events located in space \cite{nuchter2008towards}. An early work proposes concurrent object identification and localization using a supervised hierarchical neural network classifier on image color histogram feature vectors \cite{kestler2000concurrent}. However, because these approaches rely on supervised datasets \cite{genova2021learning,vineet2015incremental}, they work only on a closed set of vocabularies and do not generalize to open-ended semantic queries. 

More recent works have focused on using large vision-language models to support open-vocabulary queries. This includes both 2D~\cite{brohan2023can,brohan2022rt,zitkovich2023rt} and 3D, such as VL-Maps~\cite{huang23vlmaps} and CLIP-fields~\cite{shafiullah2023clipfields}, which assigns a CLIP feature to every point in the 3D scene. This can be used for setting navigation goals with natural language queries. OpenScene~\cite{peng2023openscene} ensembles open-vocabulary feature encoders and 3D point networks to form a per-point feature-vector allowing natural language querying on pointclouds. ConceptFusion~\cite{jatavallabhula2023conceptfusion} develops 3D open-set multimodal mapping by projecting CLIP~\cite{radford2021learning} and pixel aligned features into 3D points, and additionally fuse other modalities such as audio into the scene representation. 
ConceptGraphs~\cite{conceptgraphs} model spatial relationships as well as the semantic objects in the scene to reason over spatial and semantic concepts. 

Semantic fields have been applied not only to scene-level understanding for mobile robots but also to manipulation. In these settings, NeRFs~\cite{mildenhall2021nerf} have been a popular 3D representation, following from Distilled Feature Fields~\cite{kobayashi2022decomposing}, Neural Feature Fusion Fields~\cite{tschernezki2022neural}, and Language Embedded Radiance Fields (LERFs)~\cite{kerr2023lerf}. These works learn an semantic field in addition to the color field. LERF supports a scale-conditioned feature field, which takes in an extra scalar as input to facilitate feature encodings at multiple scene scales. For manipulation, the feature fields have been shown to facilitate learning from few-shot demonstrations~\cite{shen2023F3RM}, policy learning~\cite{ze2023gnfactor}, zero-shot trajectory generation~\cite{huang2023voxposer}, and task-oriented grasping~\cite{lerftogo2023}. 

LERF-TOGO's~\cite{lerftogo2023} zero-shot task-oriented grasping performance is fully based on LERF, as LERF's multi-scale semantics allow for both object- and part-level understanding. This property is also valuable in scene-level settings, where a human may specify a collection of objects, e.g., utensils. 
LEGS maintains this multi-scale understanding, while speeding up training and querying time, by using Gaussian Splats~\cite{kerbl20233d} which have a significantly faster render time. 

\subsection{3D Gaussian Splatting}
3D Gaussian Splatting (3DGS) originated in 2023~\cite{kerbl20233d} to model a scene as an explicit collection of 3D Gaussians. Each Gaussian is described by its position vector $\mu$, covariance matrix $\Sigma$, and an opacity parameter $\alpha$, creating a representation that is both succinct and adaptable for static environments. The choice of 3D Gaussians over traditional point clouds is strategic; their inherent differentiability and the ease with which they can be rasterized into 2D splats enable accelerated $\alpha$-blending during rendering. By avoiding volumetric ray casting employed by Neural Radiance Fields (NeRFs), Gaussian Splatting has a substantial speed advantage and can support real-time rendering capabilities.

Soon after its release, 3DGS has been applied to  mapping~\cite{keetha2023splatam}, semantic mapping~\cite{li2024sgsslam}, navigation~\cite{chen2024splatnav}, and semantic fields~\cite{zuo2024fmgs,qin2023langsplat}. 3DGS's fast rendering time speeds up optimization, making it suitable for integrating visual SLAM and natural language queries for 3D semantic fields. 3DGS has also been demonstrated in both indoor datasets~\cite{dai2017scannet,yeshwanth2023scannet,Schops_2019_CVPR}, and outdoor driving scenes with multiple cameras where all sensor data is collected before 3DGS training~\cite{zhou2024drivinggaussian}. 

\subsection{Concurrent Research}
Other work in 3D Gaussian Splatting has focused on embedding language features, and separately, training online. 

Learning semantic features for Gaussians have taken either one of two approaches: calculating it on-the-fly by querying a network or maintaining multi-dimensional features for each Gaussian. FMGS \cite{zuo2024fmgs} uses multi-resolution hash encodings~\cite{mueller2022instant} optimized with a render-time loss to combine CLIP features with a map of 3D Gaussians. LEGS similarly utilizes a hash encoding for its feature field, however it includes scale-conditioning as opposed to averaging CLIP across scales, retaining finer-grained language understanding. On the other hand, LangSplat \cite{qin2023langsplat} embeds language in 3DGS by training a scene-specific autoencoder to map between CLIP embeddings and a lower-dimensional latent feature associated with each 3D gaussian. Like traditional radiance field methods, LangSplat assumes poses are corresponded with all scene images prior to 3DGS training as it requires training a VAE over all images of a scene before starting its 3D optimization. However, for robotic systems, it is often desirable to develop 3D semantic understanding online as the robot explores new and previously unseen large-scale environments.

SplaTAM \cite{keetha2023splatam} optimizes both camera pose and the 3D Gaussian map simultaneously for single-camera setups. However, having multiple cameras and viewpoints can enhance efficient environment data collection. Additionally, SplaTAM lacks semantic features, which is important for identifying and interacting with objects in a 3D scene.

To our knowledge, LEGS is the first system that integrates the advantages of both online 3DGS training and language-aligned feature supervision into Gaussian splats for large-scale scene understanding.

\section{Problem Statement}


We consider a large indoor environment, specifically defined as a room encompassing at least 750 sq ft. The objective is to 3D reconstruct the a-priori unknown and unstructured environment and localize objects prompted by open-vocabulary and long-tail natural-language queries.


We make the following assumptions:
\begin{enumerate}
    \item The environment and all objects within it are static.
    \item Queried objects are seen at least from one of the cameras.
    \item A mobile robot with 3 orthogonal cameras. For the grocery store environment, the TTT robot with a single pair of stereo cameras is used \cite{bajracharya2024demonstrating}.
    
\end{enumerate}

For each trial, the system is prompted by a natural-language query, and outputs the heatmap and localized 3D coordinate of the most semantically relevant location in the scene. The trial is deemed successful if this point falls within a manually annotated bounding box for that object. The objective is to efficiently build a 3D representation that maximizes this success rate for large-scale scenes.

\section{Methods}

We use a Fetch mobile robot equipped with an RGB-D Realsense D455 camera and two side facing ZED 2 stereo cameras mounted with known relative poses. We build a map of the scene with a set of 3D Gaussians in an online fashion, registering new images as the robot drives around the environment. The overall pipeline for LEGS is outlined in Figure \ref{fig:system_integration}. There are three key components of the system:
\begin{enumerate}
\item \textbf{Multi-Camera Reconstruction}: 
To improve the effective field-of-view of the robotic system, we use multiple cameras pointing in different directions to provide more viewpoints of the environment.
\item \textbf{Incremental 3DGS Construction}: A significant challenge of large scene mapping is localization error due to its accumulative nature. To mitigate this error, we perform global Bundle Adjustment (BA) with DROID-SLAM to improve pose accuracy for all previously recorded poses in the scene. Global BA can be executed multiple times in a given traversal, and after each BA, the prior image-pose estimates are updated with the corresponding pose.

\item \textbf{Language-Embedded Gaussian Splatting}: We implement a language-aligned feature field inspired by the method from LERF~\cite{kerr2023lerf} that samples from gaussian primitives instead of from a density field.


\end{enumerate}
\subsection{Multi-Camera Reconstruction}
Online image registration enables Gaussian Splatting on a mobile base with a comprehensive sensor suite including multiple camera views (left, right, center). During testing of vanilla Gaussian Splatting on our multi-camera setup, we found that offline Structure from Motion (SfM) pipelines frequently failed to find correspondences between images from different cameras, largely due to the lack of scene overlap from offset camera views. Because we peform online image registration with a visual SLAM algorithm for one camera and we know the corresponding extrinsic transforms to the other cameras, we can compute the corresponding pose estimate for each camera. Due to limited GPU memory onboard the Fetch robot, the image data is then streamed over network to a desktop computer for processing and training.


\subsection{Incremental 3DGS Construction + Bundle Adjustment}
Standard methods for radiance field optimization require image-pose pairs as input, and Gaussian Splatting greatly benefits from having a pointcloud as a geometric prior for initialization. Poses and pointclouds are typically provided through \textit{offline} Structure from Motion (SfM) techniques like COLMAP~\cite{agarwal2011building} which require all training images to be collected ahead of time. We build off of Nerfstudio's~\cite{tancik2023nerfstudio} Splatfacto implementation of Gaussian Splatting and modify it to operate on a stream of images and poses, and incorporate updates from global bundle adjustment (BA) to further optimize poses by building a keyframe graph to minimize the Mahalanobis distance between the reprojected points and the corresponding revised optical flow points \cite{teed2022droidslam}.
\subsubsection{Online Optimization}
For online pose estimation we use
DROID-SLAM~\cite{teed2022droidslam}, a monocular SLAM method that takes in monocular, stereo, or RGB-D ordered images and outputs per-keyframe pose estimates and disparity maps. 
During operation, we feed DROID-SLAM input frames from one of the side-facing Zed cameras, and extrapolate the poses of other cameras using the camera mount CAD model. Registered RGBD keyframes from DROID-SLAM are incrementally added to Splatfacto's training set. We initialize new Gaussian means per-image by sampling 500 pixels from each depth image and deproject them into 3D using the corresponding metric depth measurement.
We use a learned stereo depth model trained on synthetic images~\cite{shankar2022learned}, which can predict thin features and transparent objects with high accuracy. 

\subsubsection{Global Bundle Adjustment}
\begin{figure*}[!t]
    \centering
    \includegraphics[width=0.999\linewidth]{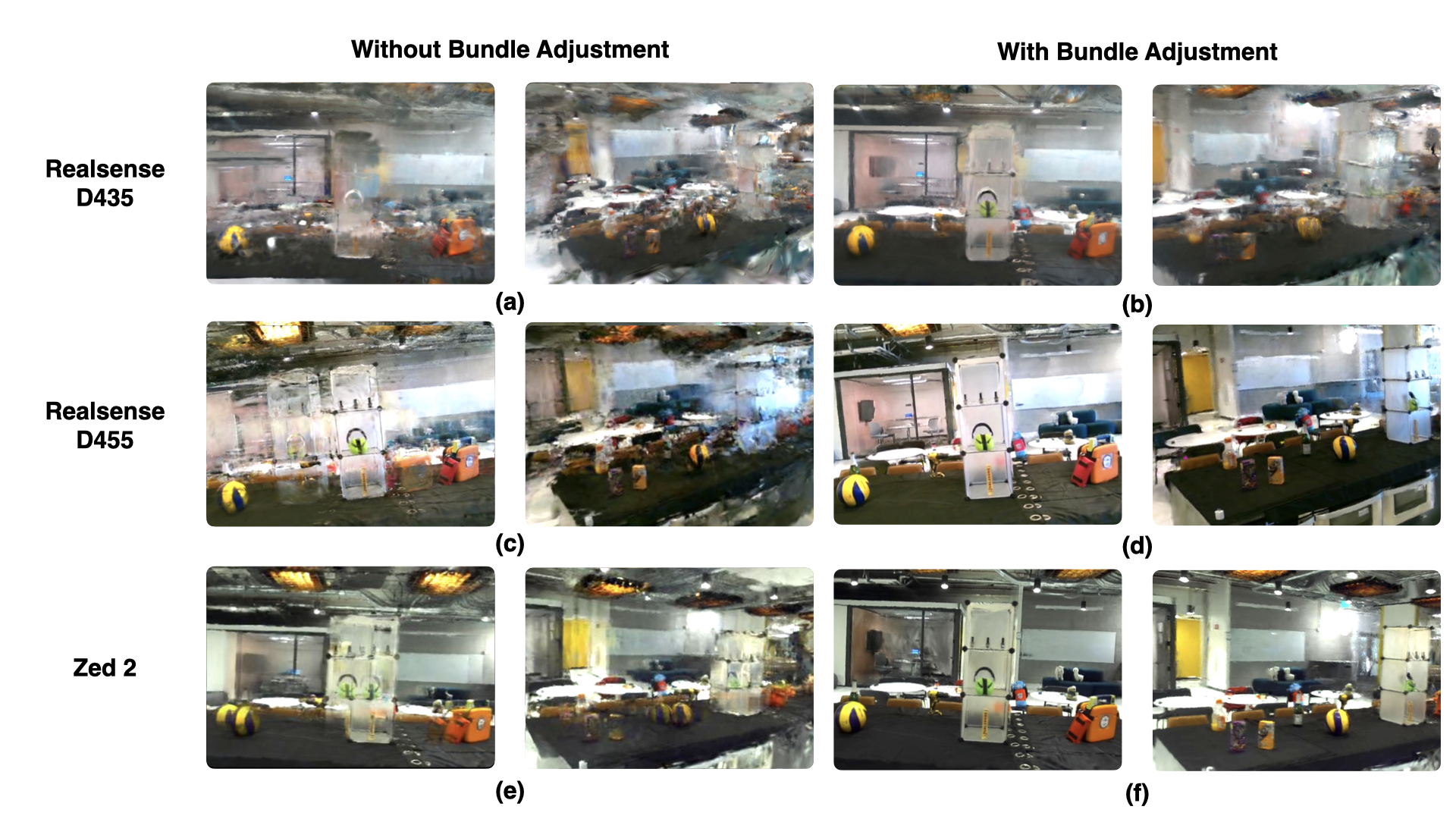}
    \caption{\textbf{Single Camera Reconstruction Comparison Results.} We compare the quality of Gaussian splats on an Intel Realsense D435, Intel Realsense D455, and Stereolabs Zed 2 with and without bundle adjustment. For each configuration we present two views: one of the Gaussian splat facing the kitchen island head-on and another view at an angle.}
    \label{fig:sensitivity_analysis}
    \vspace*{-0.3in}
\end{figure*}

Incorporating images with pose drift from SLAM systems results in artifacts in 3DGS models like duplicated or fused objects, fuzzy geometry, and ghosting (Fig.~\ref{fig:sensitivity_analysis}). Though prior work has demonstrated that pose optimization inside a 3DGS can track camera pose~\cite{keetha2023splatam}, tracking iterations for the method with reported results takes up to a second to perform, making it difficult for online usage. Instead, to mitigate drift, we incorporate updates from global BA and update training camera poses in the 3DGS accordingly. This allows tracking of new camera frames at 30fps in tandem with continual 3DGS optimization for faster model convergence.

\begin{figure}[t]
    \centering
    \includegraphics[width=0.9\linewidth]{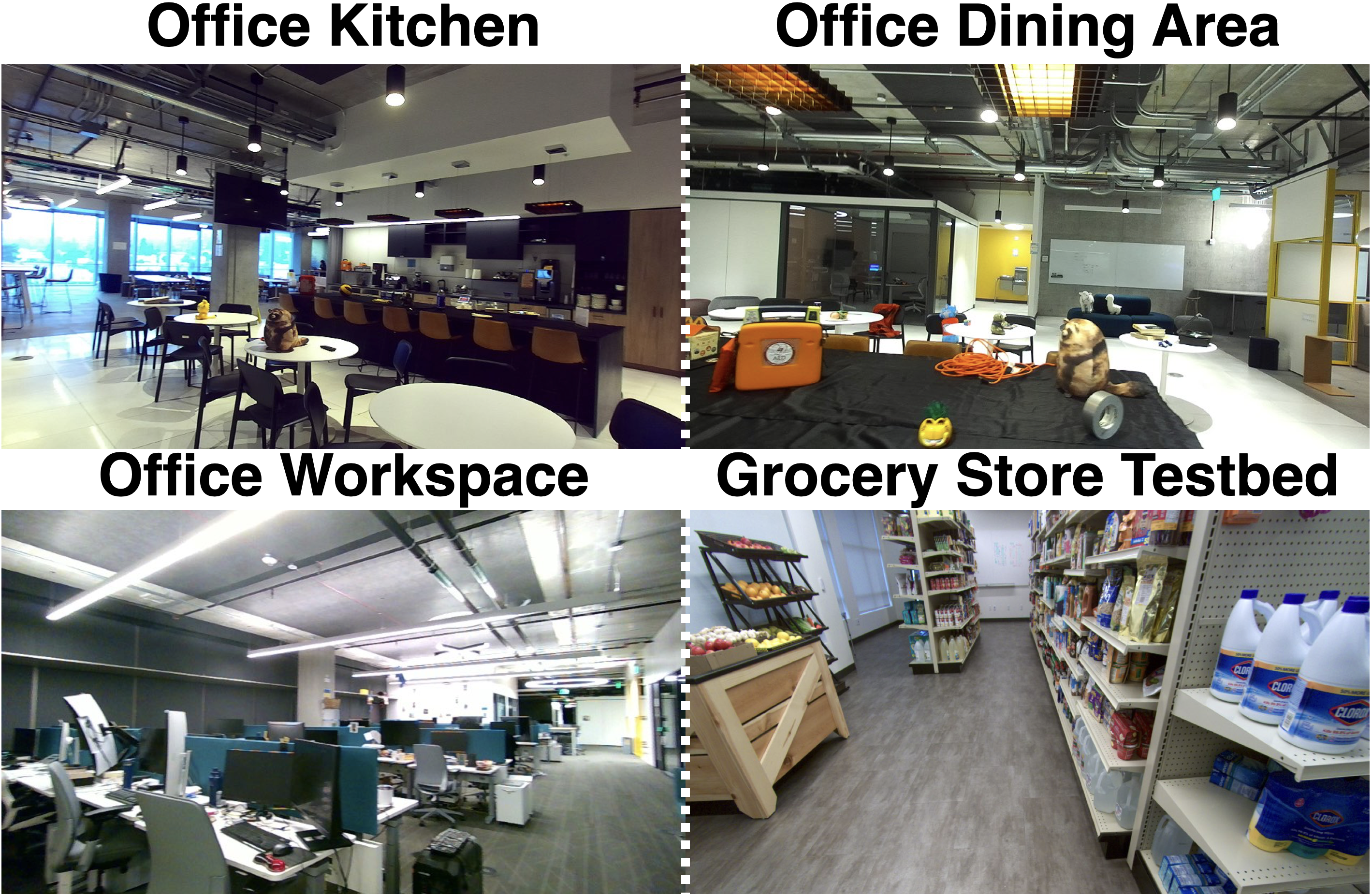}
    \caption{\textbf{4 Scene Environments.}}
    \label{fig:environments}
    \vspace*{-0.3in}
\end{figure}

\subsection{Language Embedded Gaussian Splats}
In Language Embedded Radiance Fields (LERF)~\cite{kerr2023lerf} the language field is optimized by volumetrically rendering and supervising CLIP embeddings along rays during training. In contrast, 3DGS provides direct access to explicit gaussian means, allowing us to implement a multi-scale language embedding function, $F_{lang}(\vec{x}, s) \in \mathbb{R}^D$. This function takes an input position $\vec{x}$ and physical scale $s$, outputting a $D$-dimensional language embedding. $F_{lang}$ is implemented by passing a sampled $\vec{x}$ through a multi-resolution hash encoding~\cite{mueller2022instant}, which produces the input $z$ to the MLP $m_\theta(z, s)$, with the MLP evaluated last resulting in feature output $y \in \mathbb{R}^D$. These features can then be projected and rasterized into feature images using Nerfstudio's~\cite{tancik2023nerfstudio} tile-based rasterizer implementation, with loss gradients backpropagated through the MLP. Hash encoding employs a hash table to store feature vectors corresponding to grid cells in a multi-resolution fashion. This hash grid encoding scheme effectively reduces the number of floating-point operations and memory accesses needed during training and inference, compared to directly feeding position features into an MLP.

Language-aligned features are obtained from the training images using multi-scale crops passed through the CLIP encoder, a technique shown in LERF to be crucial for semantic understanding in large scenes where object sizes may vary drastically. This is in contrast to prior work which averages CLIP embeddings as a tradeoff between speed and accuracy~\cite{zuo2024fmgs}. LEGS facilitates inference at approximately 50 Hz 1080p, and its hybrid explicit-implicit representation allows faster scene querying without volumetric rendering.

Given a natural language query, we query the language field to obtain a relevancy map similar to LERF. To localize the query in the world frame, we find the relevancy over CLIP features and take the argmax relevancy over 3D gaussian means. This method offers a significant speed increase over feature field methods distilled in NeRFs, where the volumetric representation must first render a dense pointcloud.

\section{Experiments}

\subsection{Physical Experiments}
We evaluate LEGS through a series of open-vocabulary object recall tasks. These tasks are designed to measure the system's competency in capturing and organizing information based on both location and semantic meaning. We evaluate LEGS on four large-scale indoor environments, two office kitchen scenes containing different objects, an office workspace, and a grocery store testbed \cite{bajracharya2024demonstrating} as seen in Figure \ref{fig:environments}. For the grocery store testbed, data is collected with the TTT robot \cite{bajracharya2024demonstrating}. The robot begins in a previously unseen environment and is manually pushed around a pre-planned path (including straight lines, loops, figure-8, etc) while continuously registering new images until it finishes the path. The robot will actuate its torso height to obtain multiple azimuthal perspectives from the same position on subsequent passes. Every 150 keyframes, we perform global bundle adjustment on all previous poses in DROID-SLAM and update accordingly in our 3D Gaussian map. Our system uses 2 NVIDIA 4090s, one for training LEGS, which takes 15 GB of memory and the other for DROID-SLAM, which can take up to 18 GB of memory.

 The evaluation approach was adopted from previous 3D language mapping works \cite{kerr2023lerf}, \cite{shi2023language}. We randomly sample images from our training set and query Chat-GPT 4V with: ``name an object in this scene in 3 words or fewer". This process is repeated until 15 unique queries are generated for each of the four scenes. We then choose a random novel view and manually annotate a 2D bounding box around each selected object of interest. Then, we query LEGS on each object and identify the highest activation energy point, and project that point in the novel 2D view. If the projected point is contained in the bounding box, we consider the query successful.  Additionally, we directly baseline our approach by running LERF \cite{kerr2023lerf} to compare the object recall capabilities for a large-scale scene in radiance field methods. 

\begin{figure}[!t]
    \centering
    \includegraphics[width=0.70\linewidth]{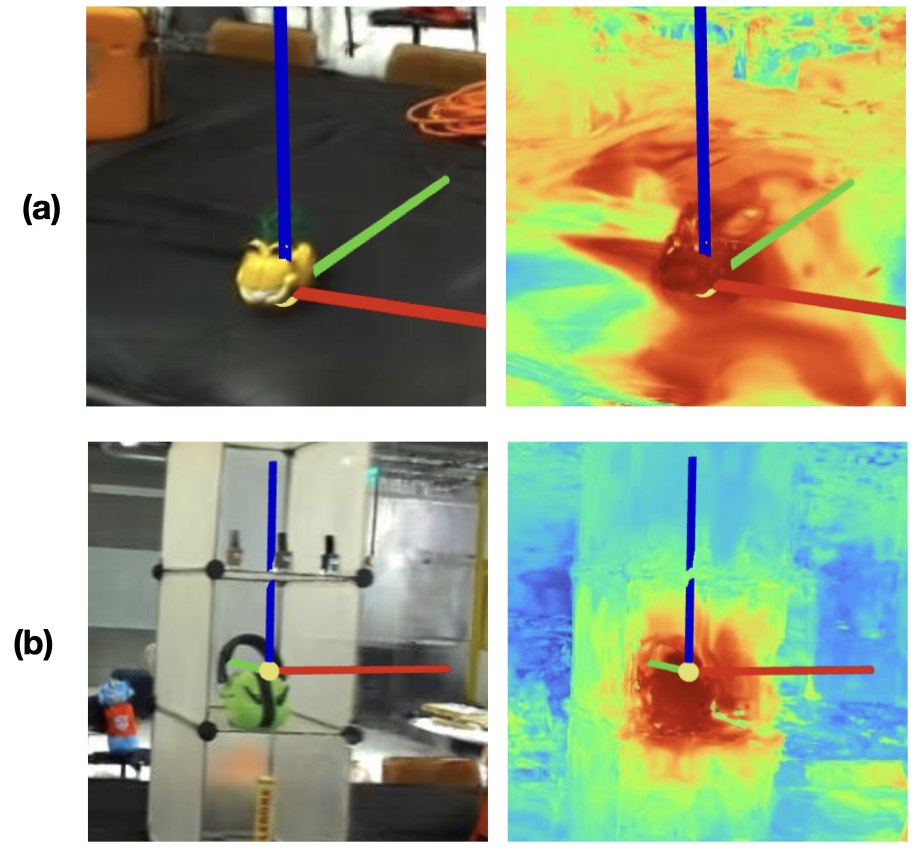}
    \caption{\textbf{Successful query localization results.} Coordinate frames on open-vocabulary and long-tail objects (a) ``garfield," (b) ``hearing protection."}
    \label{fig:successful_queries}
    \vspace*{-0.2in}
\end{figure}

\begin{figure}[!t]
    \centering
    \includegraphics[width=0.70\linewidth]{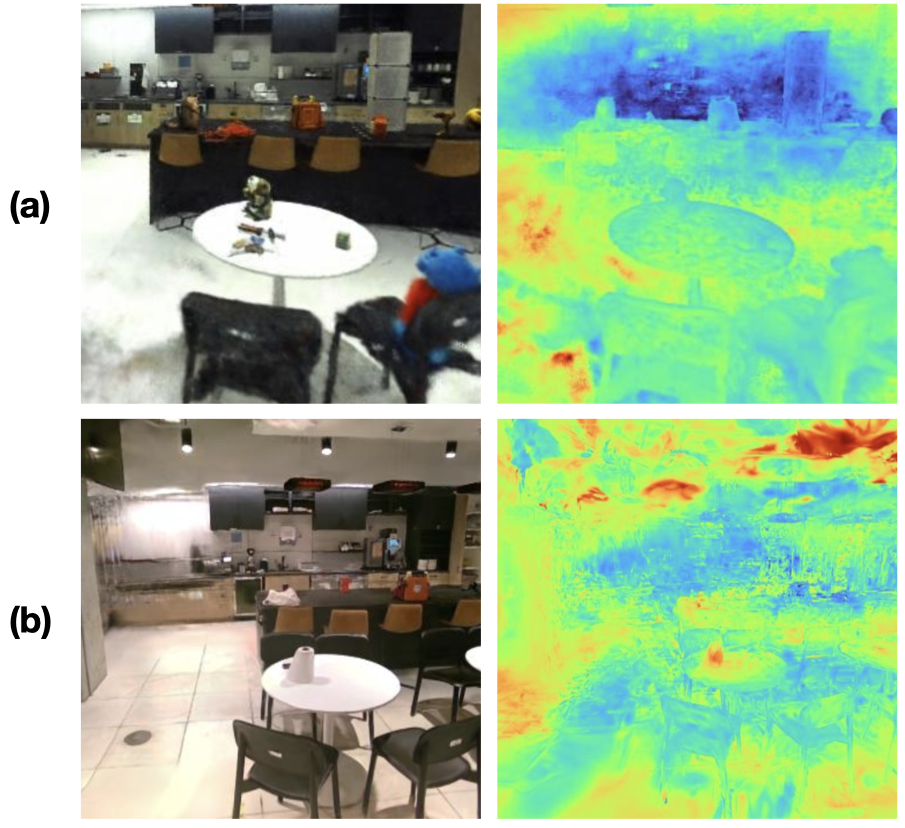}
    \caption{\textbf{Failed query localization} on (a) ``scissors," (b) ``paper roll." Where object of interest is too small in the training view or lack of distinctive color features.}
    \label{fig:fail_queries}
    \vspace*{-0.3in}
\end{figure}

The results in Table 1 suggest that LERF and LEGS have similar language capabilities, recalling roughly the same number of objects per scene. However, to achieve the same visual quality, LERF takes an average time of 44 minutes to train while LEGS only takes 12 minutes. Figure \ref{fig:successful_queries} shows examples of successful object localization queries.
Localization may fail when objects are not seen well in the training views or have similar color to their background, as shown in Figure~\ref{fig:fail_queries}.

\begin{table}
\vspace{13pt}
  \centering
  \begin{tabular}{c c c c}
    \hline 
    \textbf{Environment} & & \textbf{LERF} & \textbf{LEGS} \\
    \hline
    Office Kitchen & & 9/15 & \textbf{10/15} \\
    Office Dining Area & &  \textbf{11/15} & \textbf{11/15} \\
    Office Workspace & & \textbf{10/15} & 9/15 \\
    Grocery Store Testbed & & \textbf{12/15} & 10/15 \\
    \hline
    Avg. Train Time & &  44min. & \textbf{12min.}
  \end{tabular}

\label{tab:opt:sample}
\caption{\textbf{Object recall success rates.} Comparison between LERF and LEGS on large scenes where both receive the same SLAM poses. LEGS receives poses incrementally, while LERF receives the final poses. Average train time refers to the time until 20 average PSNR. For LEGS we consider the time after the final image is added to the train set.}
\vspace*{-0.1in}
\end{table}


\subsection{Reconstruction quality comparison}

\begin{table}
  \centering
  \begin{tabular}{c c c c}
    \hline 
    & {\phantom{a}} & \multicolumn{2}{c}{\textbf{PSNR}} \\
 
    \textbf{Camera} & & \textbf{w/o BA} & \textbf{w/ BA} \\
    \hline
    D435 & & 18.6 & \textbf{22.7} \\
    D455 & & 20.0 & \textbf{23.5} \\
    Zed 2 & & 19.2 & \textbf{23.8} \\
    \hline
  \end{tabular}
\caption{\textbf{PSNR (Peak Signal-Noise Ratio) scores} across different cameras with and without bundle adjustment quantifying training-view reconstruction quality. Trained until 20k iterations after final image is added to the train set.}
\label{tab:psnr}
\vspace{-20pt}
\vspace*{-0.1in}
\end{table}

We study how camera configuration and bundle adjustment (BA) affect the quality of the LEGS Gaussian splat as summarized in Figure \ref{fig:sensitivity_analysis} and Table 2. With respect to the camera configuration ablation, we evaluated different depth and stereo cameras including the Realsense D435, Realsense D455, and Zed 2. For each camera configuration, we also run a BA ablation where we either run bundle adjustment at the end of the traversal or not at all. 

\textbf{Global Bundle Adjustment}: Table~\ref{tab:psnr} suggests bundle adjustment improves Gaussian Splat quality for all camera configurations, removing ghostly duplicate artifacts. This is especially true for the Realsense D455 and Zed 2 cameras where the bundle adjustment configurations yielded near-photorealistic views of the scene whereas without bundle adjustment, both configurations have significantly more Gaussian floaters and/or offset objects (i.e. the left image in Figure \ref{fig:sensitivity_analysis} part (e) has two volleyballs). The Realsense D435 performs slightly better with bundle adjustment, but neither D435 configuration yield high quality largely due to the camera's low FOV resulting in worse localization.

\begin{figure}[t]
    \centering
    \includegraphics[width=0.9\linewidth]{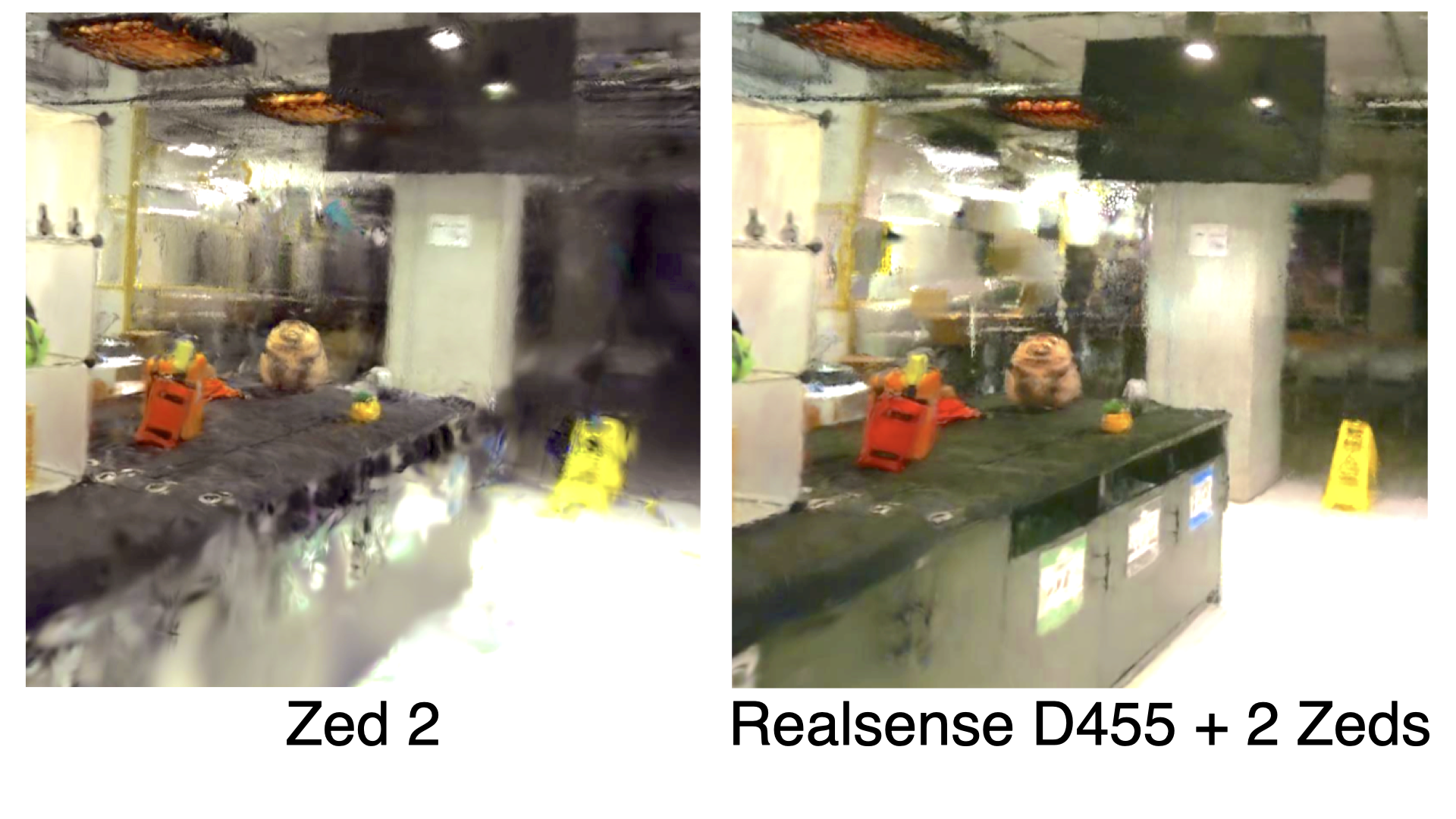}
    \caption{\textbf{Single Camera vs Multi Camera Reconstruction.} With the multi-camera setup, the effective field of view is increased, elucidating more of the scene such as the wet floor sign and trash chutes on the side faces of the table.}
    \label{fig:multi_cam_sensitivity_analysis}
    \vspace*{-0.3in}
\end{figure}

We also compared a single Zed 2 camera to a multi-camera setup where the D455 Realsense is front-facing and 2 Zed cameras face the left and right side. Both gaussian splats perform well and properly render objects that were well viewed in the traversal (``raccoon toy" and ``first aid kit") as seen in Figure \ref{fig:multi_cam_sensitivity_analysis}. However, none of the cameras were pointing toward the ground leading to sparse views of objects near the floor. Because the multi-camera setup captures more views of the scene, it is able to construct a Gaussian splat that is better able to render these low-view objects such as the trash chutes and wet floor sign.


\section{Limitations}

We assume a static environment where objects do not move during traversal. This limits the scope of this work because many applications involve dynamic scenes with moving objects. 
In future work, we will adapt our method to work for dynamic scenes.

The motion of the Fetch mobile base can have a large effect on the LEGS reconstruction quality; the high stiction between the robot's caster wheels and the environment introduces jolts, causing camera pose inaccuracies and image blurs. 
In the future, we hope to correct this with a new mobile base where the trajectory is autonomously determined by a frontier-based exploration algorithm \cite{topiwala2018frontier}.


Although autonomous navigation and obstacle avoidance has been extensively studied \cite{pandey2017mobile}, obstacles can pose a problem when it comes to the 3D Gaussian map if they are only visible in a few of the ground truth images. 3D Gaussians are initialized at the deprojected points from these few images, but there are not enough views to refine and properly train these Gaussians; the result is oddly colored floaters that obstruct some parts of the static scene.

When performing natural language queries, LEGS inherits the limitations of LERF + CLIP distillation into 3D described by similar works \cite{kerr2023lerf}. In our experimentation, we find that a large scale environment brings additional challenges in querying, particularly in 1) small or far-field objects in the training view, 2) similar item-background color features, such as white objects on white. Language embedded Gaussian splats can also produce false-positives when querying an object that is not in the scene due to the presence of visually or semantically similar objects, which may get incorrectly classified as the query object.
\section{Conclusion} 
\label{sec:conclusion}

In this work, we introduce Language-Embedded Gaussian Splats (LEGS), a system that can train Gaussian Splats online with CLIP embeddings for large-scale indoor scenes. Because of pose accumulation error that builds up in large scenes, we use incremental bundle adjustment to improve pose fidelity for Gaussian Splat training. Results suggest LEGS trains 3.5x faster than LERF with comparable object recall.

\renewcommand*{\bibfont}{\footnotesize}
\printbibliography
\end{document}